\documentclass[twocolumn,5p]{elsarticle}  
\usepackage{setspace}  

\usepackage{amsfonts}
\usepackage{multirow}
\usepackage{eurosym}

\usepackage{graphicx, graphics}
\newtheorem{theorem}{Theorem}

\newtheorem{algorithm}[theorem]{Algorithm}

\usepackage{hyperref}

\biboptions{numbers,sort&compress}
\usepackage{algorithm}
\usepackage{algorithmicx}
\usepackage{algpseudocode}
\usepackage{amsmath} 
\usepackage{booktabs}
\usepackage{optidef}

\floatname{algorithm}{Algorithm}

\begin{document}
	\begin{frontmatter}		
		\title{Handwritten and Printed Text Segmentation via Region-aware Human-writing  Descriptor  Engineering}		
		\author[a,b]{Zhixian Lu}
		\affiliation[a]{
			organization={College of Computer Science, Chengdu University},
			addressline={No. 2025, Chengluo Avenue, Longquanyi District},
			postcode={610106},
			city={Chengdu},
			country={China}
		}
		\affiliation[b]{
			organization={Key Laboratory of Digital Innovation of Tianfu Culture, Sichuan Provincial Department of Culture and Tourism, Chengdu University},
			city={Chengdu},
			country={China}
		}
		
		\author[a,b]{Jianwei Zhang\corref{corb}}
		\ead{zhangjianwei@cdu.edu.cn}    
		
		\author[a,c,d]{Lei Zhang}
		\affiliation[c]{
			organization={Machine Intelligence Laboratory, College of Computer Science, Sichuan University},
			city={Chengdu},
			country={China}
		}
		\affiliation[d]{
			organization={Tianfu Jincheng Laboratory},
			city={Chengdu},
			country={China}
		}
		
		\author[e]{Fei Yuan}
		\affiliation[e]{
			organization={Network Information Centre, Hainan College of Economics and Business},
			city={Haikou},
			country={China}
		} 
		
		\author[a]{Jin Wang}
		\author[a]{Chang Liu}
		\author[a]{Rui Gao}
		\author[a]{Qiyu Lei}
		
		\cortext[corb]{Corresponding author}
		
		\begin{abstract}
			With the increasing demand for reusing paper documents in educational and office settings, accurate segmentation of handwritten and printed text has become a crucial step in document digitization. Although numerous deep learning models have been developed for this task, their high computational cost limits deployment on resource-constrained edge devices. To address this challenge, we present a lightweight framework optimized for efficient performance on devices with severely limited computational capacity. Our approach begins with the Sentence-level Connected Component Segmentation algorithm, aimed at extracting coherent sentence-level segments from document images. We then design a novel Region-aware Handwriting Descriptor (RHD) to capture the intrinsic variability of human handwriting at the sentence level. A simple conventional classifier can then be seamlessly integrated with our designed descriptor, demonstrating strong classification performance for distinguishing handwritten and printed sentence-level text images, highlighting that the proposed descriptor is agnostic to the choice of classifier. Extensive experiments are performed on our self-constructed Multilingual High-Quality Annotated Dataset for Handwritten and Printed Text Segmentation~(MAD-HPTS) and a public benchmark PHD-AS, and the experimental results demonstrate that the proposed framework outperforms current state-of-the-art methods in both accuracy and computational efficiency. On MAD-HPTS, our method sacrifices only $1.4\%$ accuracy compared to the leading deep neural network baseline, yet achieves more than $8\times$ speedup in inference, making it well-suited for lightweight deployment.
		\end{abstract}
		
	\end{frontmatter}

	\section{ Introduction }
	The widespread use of paper documents in educational and office environments necessitates efficient methods for their digital transformation. Among the various processing steps, Handwritten and Printed Text Segmentation~(HPTS) remains a fundamental challenge in document digitization~\cite{kuhnke1995system, pal1999automatic}. This task underpins a range of downstream applications, including document retrieval~\cite{srivastva2013survey}, signature verification~\cite{shirdhonkar2010discrimination}, and Optical Character Recognition~(OCR) enhancement~\cite{belaid2013handwritten}. For example, HPTS can dramatically accelerate document retrieval compared to manual inspection, and it plays a pivotal role in financial systems, where accurate handwritten signature extraction directly impacts security and privacy~\cite{srivastva2013survey, belaid2013handwritten, shirdhonkar2010discrimination}. In addition, recent studies report that isolating handwritten content can improve OCR accuracy by up to more than $20\%$~\cite{jo2020handwritten, vafaie2023improvements}.
	
	\begin{figure*}[htb]
		\centering
		\includegraphics[width=\linewidth]{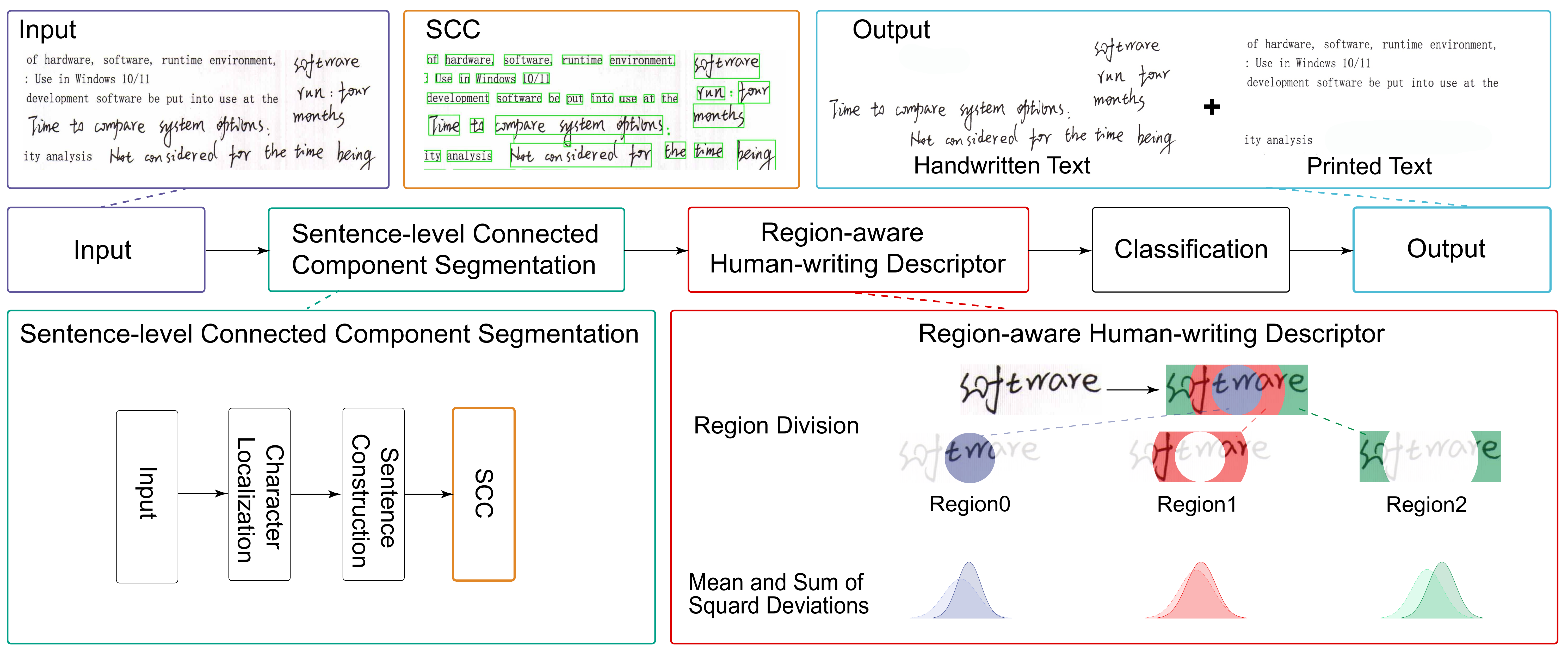}
		\caption{The proposed framework processes an input image by first applying the SCCS algorithm to segment it into Sentence-level Connected Components~(SCCs). For each SCC, we extract the RHD feature and classify it as handwritten or printed text.}
		\label{fig:architecture}
	\end{figure*}
	
	Historically, HPTS methods can be broadly categorized into two main classes: (1)~machine learning-based methods~\cite{zheng2001single, zheng2002segmentation, peng2009markov, kavallieratou2004machine, malakar2013handwritten, garlapati2017system}, and (2)~deep learning-based methods~\cite{vafaie2023improvements, jo2020handwritten, dutly2019phti}. Early research predominantly focused on machine learning-based methods, which typically follow a three-stage pipeline: (i)~extracting connected components~(CCs)—such as characters, words, or sentences—from the document image; (ii)~deriving features from these components using texture-based descriptors such as Gabor filters~\cite{zheng2002segmentation} or Local Binary Patterns (LBP)~\cite{malakar2013handwritten, garlapati2017system}; and (iii)~classifying the CCs based on the extracted features using conventional machine learning classifiers. These methods offer fast feature extraction owing to their lightweight designs. However, their performance heavily depends on handcrafted features, which often fail to capture the complex visual variability between handwritten and printed text, thereby limiting segmentation accuracy.
	
	To overcome the limitations of machine learning-based methods, deep learning-based methods have been introduced to enable automatic feature extraction and improve the resulting segmentation accuracy~\cite{jo2020handwritten, dutly2019phti, vafaie2023improvements}. These methods typically formulate HPTS as a semantic segmentation problem and address it within an end-to-end convolutional neural network(CNN) framework. For instance, Jo et al.\cite{jo2020handwritten} employed a U-Net architecture to perform pixel-level classification of handwritten and printed text. Dutly et al.\cite{dutly2019phti} utilized a Fully Convolutional Network~(FCN) for initial segmentation, followed by Conditional Random Fields to refine isolated or ambiguous predictions via contextual inference. To address the intertwined nature of handwritten and printed text, Gholamian et al.~\cite{gholamian2023handwritten} proposed a fine-grained feature path within U-Net to enhance the extraction of low-level features. While deep learning-based methods have demonstrated substantially higher segmentation accuracy than traditional machine learning-based approaches, their reliance on large-scale computation and specialized hardware often hinders deployment on edge devices with limited processing capacity.
	
	Overall, while machine learning-based methods are computationally efficient and well-suited for deployment on resource-constrained devices, they often suffer from limited accuracy. In contrast, deep learning-based approaches deliver superior segmentation performance but require substantial computational resources, making them impractical for resource-constrained environments. This raises a key question: \textbf{Can we achieve both high effectiveness and efficiency in HPTS?}
	
	To address this issue, we propose a novel framework that substantially surpasses traditional machine learning-based approaches in performance and markedly outpaces deep learning-based approaches in speed, enabling real-time deployment on edge devices. The framework begins with an Sentence-level Connected Component Segmentation~(SCCS) algorithm to extract sentence-level connected components~(see Fig.~\ref{fig:architecture}). We then introduce a novel feature engineering method, termed the Region-aware Handwriting Descriptor~(RHD), which is specifically designed to facilitate the efficient and effective capture of the intrinsic characteristics of human handwriting. By computing statistical features from distinct regions within each sentence, RHD provides a powerful and interpretable representation that enables robust differentiation between handwritten and printed text. Notably, we find that using only simple statistics such as the mean and standard deviation can achieve performance comparable to deep learning-based automatic feature extraction, while incurring minimal computational overhead. To facilitate evaluation, we constructed the \textbf{M}ultilingual High-Quality \textbf{A}nnotated \textbf{D}ataset for \textbf{H}andwritten and \textbf{P}rinted \textbf{T}ext \textbf{S}egmentation~(\textbf{MAD-HPTS}), which contains English, Chinese, and Japanese samples, serving as a strong and comprehensive benchmark for HPTS. Extensive experiments on both the self-constructed MAD-HPTS and a public dataset PHD-AS demonstrate that the proposed framework consistently delivers competitive segmentation accuracy while maintaining high efficiency. On the MAD-HPTS dataset, the proposed framework achieves overall performance only $1.4\%$ lower than the baseline FCN, yet completes inference in less than $1/8$ of the time. When deployed on the RK3576 edge device, it attains the highest accuracy—approximately $2\%$ higher than the next-best method—while reducing runtime by about $45\%$, underscoring its suitability for lightweight, resource-constrained applications.
	
	The primary contributions of this paper are summarized as follows:
	\begin{enumerate}
		\item We construct MAD-HPTS, a multilingual high-quality annotated dataset for handwritten and printed text segmentation with sentence-level annotations in English, Chinese, and Japanese, covering diverse real-world scenarios such as multiple authors, pen types, scanning devices, and document sources. Compared to existing datasets, MAD-HPTS offers a strong and comprehensive benchmark for evaluating HPTS task.
		
		\item A novel Region-aware Human-writing Descriptor is proposed to capture handwriting variability at the sentence level. RHD introduces only negligible computational overhead yet delivers significant performance gains. Its region-aware feature modeling can be extended to other feature types, offering a promising direction for feature engineering in HPTS.
		
		\item A effective and efficiency HPTS framework is proposed to combine sentence-level connected component segmentation with RHD-based feature extraction. It achieves competitive accuracy compared to deep learning-based approaches while operating with substantially lower computational cost, making it well-suited for real-time deployment on resource-constrained edge devices.
	\end{enumerate}
	
	\section{Related Work}
	Over the past decades, research on handwritten and printed text segmentation has progressed from traditional machine learning-based approaches to modern deep learning-based approaches. In the following, we provide a review of representative methods in each category.
	
	\subsection{Machine Learning-based Methods}  
	Machine learning methods for handwritten and printed text segmentation are typically organized into three stages: preprocessing, feature extraction, and classification~\cite{srivastva2013survey, imade1993segmentation}. Among these, preprocessing and feature extraction are particularly critical to overall performance, as conventional classifiers in traditional machine learning pipelines play a relatively minor role in this task, serving primarily to leverage the extracted features. In the following, we focus our review on approaches for preprocessing and feature extraction.
	
	\subsubsection{Preprocessing}
	The goal of preprocessing is to extract Minimal Discriminative Units~(MDUs) from document images, such as pseudo-words~\cite{belaid2013handwritten}, image patches~\cite{peng2013handwritten}, or text blocks~\cite{zagoris2014distinction}, which serve as the fundamental units for subsequent classification.
	
	Existing methods can be broadly categorized as follows: (1)~Region Growing~\cite{shirdhonkar2010discrimination, chanda2010structural, guo2001separating, zheng2002segmentation, belaid2013handwritten, awal2017neighborhood, peng2013handwritten}, which begin by selecting candidate pixels according to predefined heuristics, and then progressively merge spatially adjacent regions based on morphological features or statistical rules. (2)~Profile-driven Segmentation~\cite{pal2001machine, kavallieratou2004machine, zemouri2011machine}, which analyzes vertical or horizontal projection profiles of document images to detect sparse regions. Segmentation lines are then inserted at these low-density positions, and the resulting partitions are treated as MDUs. Unlike Region Growing, this method relies on global page statistics, making it particularly effective for documents with regular layouts. 
	
	This study adopts a Region Growing-based methodology for its simplicity and robustness in transforming document images into structured text representations. Similar to Zheng et al.~\cite{zheng2002segmentation}, we also employ an iterative merging strategy that progressively combines smaller connected components into larger text structures. Unlike most existing methods, which typically select pixels, characters, or word blocks as MDUs for English documents, our method introduces the Sentence-level Connected Component as the MDU. This design avoids the excessive fragmentation caused by over-segmentation in prior methods when applied to Chinese texts with structurally complex characters~\cite{xi2002page}. Furthermore, the SCC-based preprocessing framework exhibits strong generalization ability, extending naturally to languages with different writing systems and layout styles, including English and Japanese.
	
	\subsubsection{Feature Extraction}
	After preprocessing, feature extraction is performed for each MDU to facilitate subsequent classification. The extracted features can be broadly categorized as follows: (1)~Geometric and Structural Features~\cite{pal2001machine, kavallieratou2004machine, zheng2002segmentation, zheng2004machine}: MDUs are characterized by simple geometric attributes such as height, width, area, aspect ratio, and pixel density. (2)~\textbf{Projection-based Statistical Features}~\cite{guo2001separating, zemouri2011machine, da2009automatic, farooq2006identifying, shirdhonkar2010discrimination, dhar2021hp_docpres}: These features characterize MDUs by computing histogram statistics of projected pixel values along specified directions to capture stroke density and alignment, as well as run-length features that measure transitions between foreground and background pixels to quantify continuity and structural complexity. (3)~Texture Features~\cite{zheng2002segmentation, zheng2004machine, peng2013handwritten, hangarge2013statistical, malakar2013handwritten, ghosh2021coalition}: Local textural variations are analyzed using descriptors such as Gabor filters, Local Binary Patterns (LBP), statistical measures, and pattern frequency analysis. (4)~Composite and Advanced Features~\cite{chanda2010structural, peng2013handwritten, zagoris2014distinction}: To enhance discriminative power, descriptors can be combined or replaced with advanced representations such as chain codes, Bag-of-Visual-Words, or shape-context features.
	
	Traditional approaches to handwritten and printed text segmentation often rely on generic image descriptors originally designed for general computer vision tasks, such as Gabor filters, LBP, and projection profiles. While effective at capturing local textures, these methods typically neglect the spatial distribution and channel-wise variability inherent to handwritten versus printed text. To address this limitation, we propose the Region-aware Handwriting Descriptor, which explicitly models the spatial variability of handwriting to enhance feature discriminability. The RHD is novel and computationally efficient, requiring only simple statistical features, namely the mean and sum of squared deviations~(SSD). Conceptually similar to Malakar et al.~\cite{malakar2013handwritten}, as both approaches rely on statistical descriptors, our method differs by partitioning each MDU into concentric annuli and computing statistical moments within each annulus across RGB channels. This design more effectively captures the spatial variability of handwriting and significantly improves discriminative power.
	
	\subsection{Deep Learning-based Methods}	
	With the rapid advancement of neural networks, deep learning has been applied to various scenarios to solve complex problems, such as image recognition, object detection, and medical image analysis~\cite{zhang2023sarpn, zhang2023evolutionary}. Consequently, deep learning have also been increasingly applied to handwritten and printed text segmentation. Recent studies can be broadly categorized into two groups: hybrid methods and end-to-end methods. 
	
	Hybrid approaches typically employ neural networks either as feature extractors or classifiers, which are then integrated into traditional machine learning pipelines. For example, Li et al.~\cite{li2018printed} first preprocessed document images into connected components and subsequently applied CNNs to classify them as handwritten or printed text. Dutly et al.~\cite{dutly2019phti, prikhodina2021handwritten, vafaie2023improvements} proposed a lightweight FCN-light model to perform pixel-level segmentation, followed by CRF-based postprocessing to refine the results. Although hybrid methods leverage the powerful feature extraction capabilities of neural networks and achieve improved performance over purely traditional approaches, their overall effectiveness is often constrained by pipeline complexity and error propagation across multiple integrated modules.
	
	To address these limitations, end-to-end methods formulate HPTS as a semantic segmentation problem, enabling pixel-level segmentation without relying on rule-based preprocessing or handcrafted features. Jo et al.~\cite{jo2020handwritten} employed a U-Net architecture and achieved strong performance on synthetic mixed datasets derived from IAM~\cite{marti2002iam} and PRImA~\cite{antonacopoulos2009realistic}. Similarly, Mondal et al.~\cite{mondal2020tseggan} introduced tsegGAN, a generative adversarial framework designed to disentangle and segment touching non-text components from handwritten text regions. By jointly performing segmentation and feature extraction in an end-to-end manner, these methods eliminate the modular constraints inherent to hybrid pipelines and substantially improve the robustness of HPTS. 
	
	
	
	Despite the significant performance gains achieved by previous deep learning–based methods for HPTS, their neural architectures often involve tens of millions of parameters and typically demand substantial computational resources and memory~\cite{jo2020handwritten, gholamian2023handwritten}. This factor severely restricts their applicability in resource-constrained environments. In this paper, we revisit HPTS from a machine learning perspective rather than adopting deep neural models. Although our approach does not exploit automatic feature learning of deep neural networks, we introduce a Region-aware Handwriting Descriptor that generalizes effectively to handwriting characteristics. The proposed method relies solely on simple statistical features, yet achieves performance comparable to, or even surpassing, deep learning–based counterparts, while requiring only $1/8$ of the computational time, making it particularly suitable for deployment on edge devices.  
	
	\section{Method}
	The proposed framework is illustrated in Fig.~\ref{fig:architecture} and comprises three main modules. First, the Sentence-level Connected Component Segmentation module processes the input document image into SCCs, which serve as the minimal discriminative units in this work. Second, the Region-aware Handwriting Descriptor module extracts features from each SCC to capture subtle distinctions between handwritten and printed text. Finally, a conventional classifier is applied to perform SCC classification. In this section, we focus primarily on the first two modules, as they constitute the core innovations of our approach.
	
	
	\subsection{Sentence-level Connected Component Segmentation}\label{sec:preprocessing}
\begin{figure}[!t]
		\centering
		\includegraphics[width=1.0\linewidth,height=0.5\textheight,keepaspectratio]{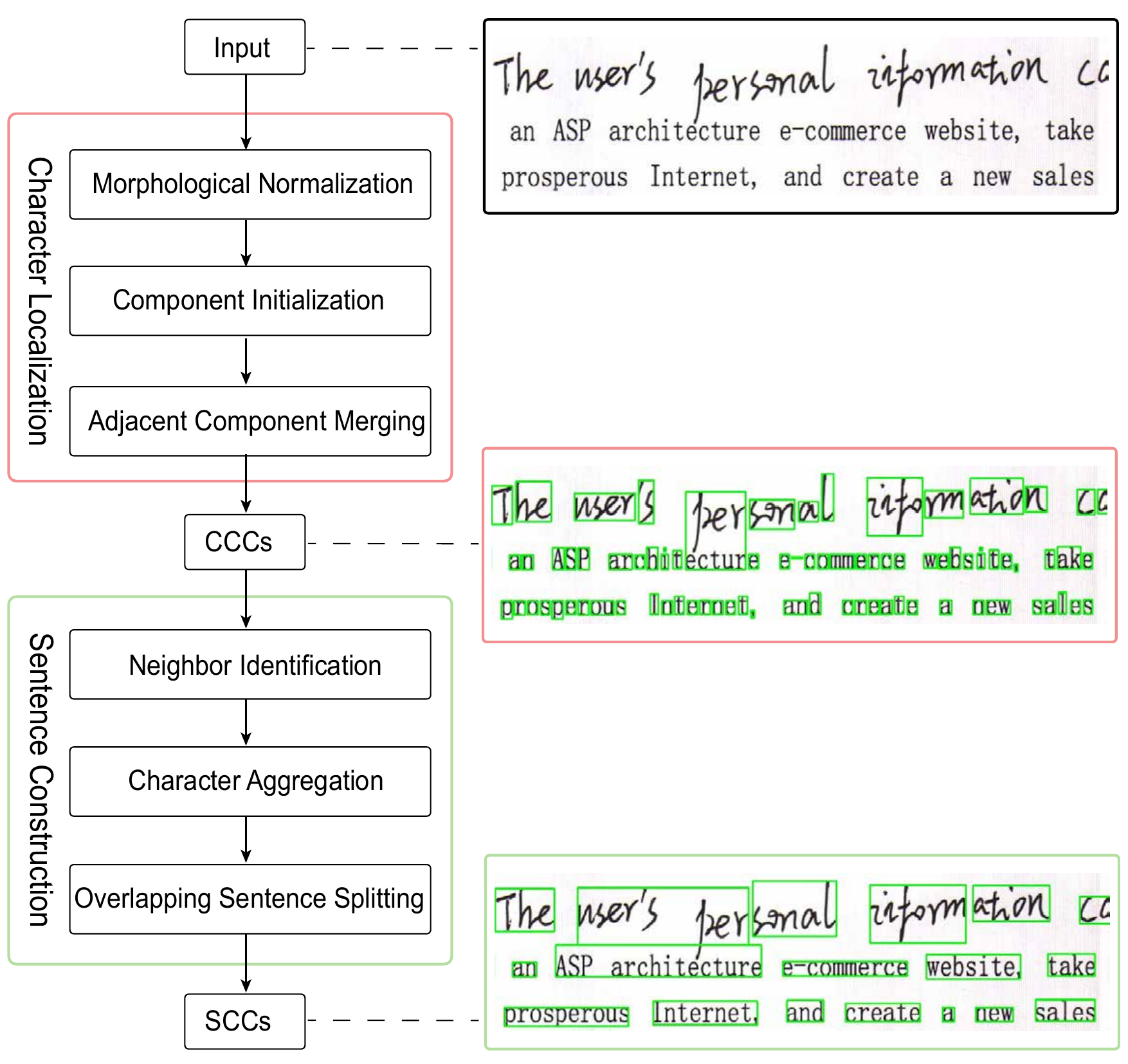}
		\caption{The pipeline illustrates the two stages of the SCCS algorithm. The input image undergoes Character Localization and Sentence Construction, generating the mask of all the Sentence-level Connected Components.}
		\label{fig:prepocessing}
	\end{figure}

	Sentence-Level Connected Component Segmentation aims to decompose the textual content of a document image into SCCs, which serves MDU in our framework. As shown in Fig~\ref{fig:prepocessing}, SCCS primarily consists of two parts: (1)~\textit{Character Localization}, which segments the input image into character-level connected components, and (2)~\textit{Sentence Construction}, which aggregates the character-level connected components into sentence-level ones in each text line. 
	
	
	\subsubsection{Character Localization}\label{CharacterLocalization} 
	Character Localization is a crucial foundation for aggregating foreground pixels into basic CCCs and the subsequent formation of SCCs. As shown in the red box in Fig~\ref{fig:prepocessing}, this process is composed of three stages: (1)~Morphological Normalization, which enhances the foreground regions and prepares them for further analysis. (2)~Component Initialization, which groups foreground pixels into basic CCs. (3)~Adjacent Component Merging, which combines nearby or overlapping CCs into more complete CCCs.  
	
	\noindent \textbf{Morphological Normalization.}  
	In this stage, the input image is first denoised using a smoothing filter to remove the background noise. In this context, foreground pixels represent text regions, while background pixels correspond to non-text areas. The Otsu thresholding~\cite{otsu1975threshold} is then applied to binarize the image, distinguishing foreground pixels from background pixels, and isolating the regions of interest. Finally, morphological erosion and dilation are applied to enhance the textual structures.
	
	\noindent \textbf{Component Initialization.}  
	A union-find algorithm is employed to merge all 4-connected adjacent foreground pixels into distinct connected components. Besides the text-related CCs (both handwritten and printed), the image also contains numerous noisy CCs that need be removed. These noisy CCs can be readily distinguished from text CCs based on differences in area, spatial density (defined as the ratio of the component area to the area of its bounding box), or aspect ratio (the length-to-width ratio of the bounding box). By applying predefined thresholds to these properties, the noisy CCs can be effectively filtered out.
	
	\noindent \textbf{Adjacent Component Merging.}  
	Many characters, such as the English letter~\textit{i} or various Chinese characters, may be split into multiple CCs. To address this, we merge any two CCs within a distance of 2 pixels into a CCC unit. As a result, the processed image is transformed into distinct handwritten and printed CCCs.
	
	\subsubsection{Sentence Construction} 
	Sentence Construction is dedicated to further transforming the character-level connected components obtained in Section~\ref{CharacterLocalization} into sentence-level connected components. As shown in the green box in Fig~\ref{fig:prepocessing}, Sentence Construction primarily consists of three stages: (1)~\textit{Neighbor Identification}, which identifies the left and right neighbors of each character; (2)~\textit{Character Aggregation}, where adjacent characters are merged into SCCs; and (3)~\textit{Overlapping Sentence Splitting}, where incorrectly merges overlapping SCCs are split into two distinct sentence-level connected components.
	
	\noindent \textbf{Neighbor Identification.}		
	For each CCC $ch$, we define its neighboring CCCs on the same text line, denoted as $\mathcal{N}(ch)= \Bigl\{\,{ch}' \in \mathcal{C} \setminus \{ch\}\;\Big|\;
	\bigl| y_{ch} - y_{{ch}'} \bigr|
	< \tau_h \,\Bigr\}$, where $\mathcal{C} \setminus \{ch\}$ represents whole CCC set $\mathcal{C}$ except $ch$ itself, $y$ denotes the vertical coordinate of the bounding box center, $h$ its height, and $\tau_h = \min \!\bigl(h_C, h_{C'}\bigr)/3$ is a tolerance parameter for filtering CCCs from other text lines. 
	
	
	\noindent \textbf{Character Aggregation.}
	Character aggregation aims to merge neighboring CCCs into SCCs, thereby forming coherent sentences. This step is particularly important in Chinese documents, as individual CCCs are often fragmented or scattered, making it difficult to support reliable classification without aggregation.
	
	\begin{algorithm}[!t]
		\caption{Character Aggregation}
		\begin{algorithmic}[1]
			\Require The complete set of CCCs $\mathcal{C}$ and the distance threshold $\tau_w$.
			\Ensure The set of all SCCs $\mathcal{S}$.
			
			\State $\mathcal{S} \gets \emptyset$ \Comment{Initialize the SCC set}
			\For{each $ch \in \mathcal{C}$}
			\State $s \gets \{ch\}$ \Comment{Initialize the current sentence}
			\While{ the character ${ch}_{end}$ at the left and right ends of the sentence $s$ has a neighbor ${ch}'$ not in $s$ \textbf{ and } their distance $D({ch}_, {ch}') \leq \tau_w$}            
			\State $s \gets s \cup \{ch'\}$ \Comment{Expand SCC by including the neighbor character}
			\EndWhile
			\State $\mathcal{S} \gets \mathcal{S} \cup \{s\}$
			\EndFor
			\State \Return $\mathcal{S}$
		\end{algorithmic}
		\label{alg:tosccs}
	\end{algorithm}
	
	\begin{figure*}[htb]
		\centering
		\includegraphics[width=\linewidth]{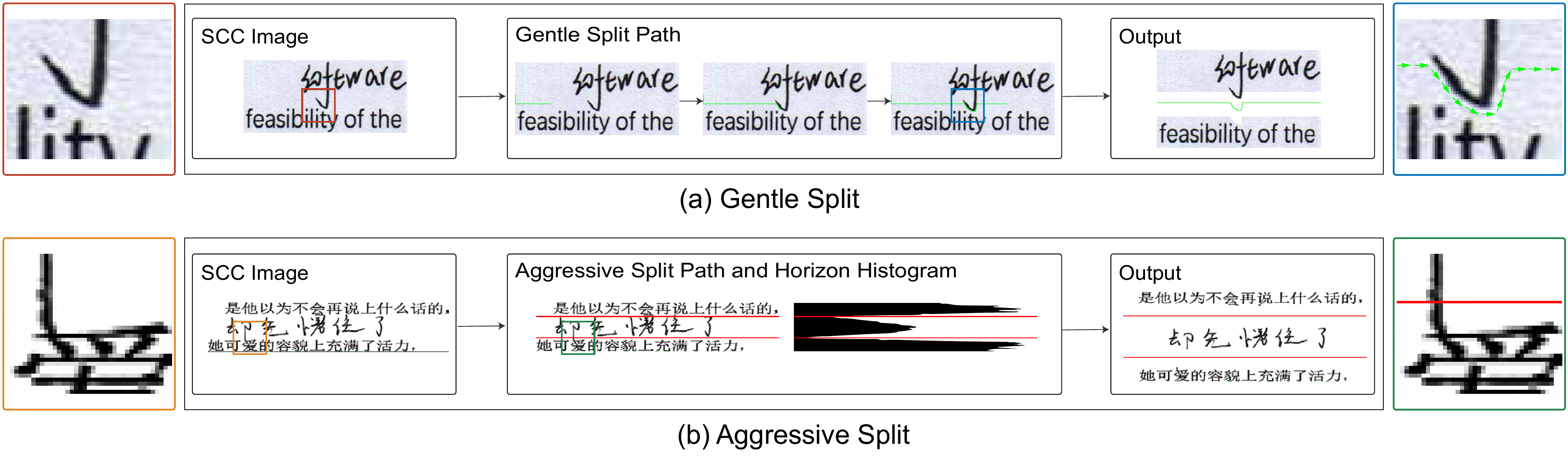}
		\caption{Illustration of the two overlapping sentence splitting strategies. (a)~\textit{Gentle Split} for oversize SCCs without actual pixel adhesion. (b)~\textit{Aggressive Split} for oversize SCCs with real pixel overlap.}
		\label{fig:split}
	\end{figure*}
	As illustrated in Algorithm~\ref{alg:tosccs}, each CCC is iteratively expanded into an SCC by merging with its left or right neighbor. Specifically, neighboring CCCs are merged when their horizontal distance is smaller than a threshold $\tau_w$, which is set as the most frequently occurring horizontal distance between neighboring CCCs.

\noindent \textbf{Overlapping Sentence Splitting.} 
As character aggregation primarily considers horizontal distances when merging characters, CCCs that are vertically misaligned but horizontally close may be erroneously merged into the same SCC, potentially resulting in overlapping SCCs that contain multiple sentences. To address this issue, we introduce an \textit{Overlapping Sentence Splitting} stage to separate such SCCs into their original text lines. In practice, two distinct types of overlapping SCCs are typically observed, and we sequentially apply two complementary strategies, \textit{Gentle Split} and \textit{Aggressive Split}, to handle each type accordingly.

\begin{algorithm}[!t]
	\caption{Gentle Split}
\begin{algorithmic}[1]
	\Require 
	The image of $SCC \in \mathcal{S}_{m}$; start pixel $(x_s, y_s)$ and target pixel $(x_t, y_t)$ pixel $(x_t, y_t)$
	\Ensure 
	A  path parent separating the SCC 
	
	\State Initialize \textbf{priority queue} $Q \gets$  ordered by $H$, Manhattan distance to target
	\State Initialize \textbf{parent} map
	\Comment{visited to record the pixel and  parent for backtracking.}
	\State $Q.\textsc{push}((x_s, y_s))$ \Comment{Initialize the start pixel}
	
	\While{$Q$ is not empty}
	\State $p \gets Q.\textsc{pop}()$ \Comment{The most closed point to target}
	\If{$p = (x_t, y_t)$}  
	\Comment{Final a path to split SCC}
	\State \textbf{break}
	\EndIf
	\For{each 4-connected neighbor $p_0$ of $p$}
	\If{$p_0$ is valid and not visited}
	
	\State Mark $p_0$ as visited
	\State Set $parent[p_0] \gets p$ \Comment{For backtracking}
	\State $Q.\textsc{push}(p_0)$
	\EndIf
	\EndFor
	\EndWhile \\
	\Return parent
\end{algorithmic}
\label{alg:gentlesplit}
\end{algorithm}

The first case corresponds to bounding-box overlap without pixel-level adhesion, as illustrated in Fig~\ref{fig:split}~(a). Here, two vertically adjacent SCCs are not truly connected, and it is possible to identify a separating path that preserves the structural integrity of both components. \textit{Gentle Split} is designed to find such a non-destructive split path, as illurstrate green pixels in the panel. As detailed in Algorithm~\ref{alg:gentlesplit}, a breadth-first search~(BFS) is performed to carve a path from the start pixel to the target pixel using a Manhattan-distance heuristic. The start and target pixels are selected near the vertical center of the left and right boundaries of the SCC, respectively. In practice, they are chosen slightly shifted toward the background-dominated side of the bounding box, preventing the path from immediately colliding with text strokes and thus facilitates a smoother segmentation. The precise offset is not critical, as the method is robust to small variations. The search is initialized with a priority queue $Q$ ordered by the Manhattan distance. The start pixel is first pushed into $Q$, and all 4-connected valid background neighbors~(Item 9-13) are considered as candidates, enqueued into $Q$, and assigned parent pointers for backtracking. And in Item 10, we define a valid pixel as a background pixel with $y$-coordinate between $\left(\frac{H}{4}, \frac{3H}{4}\right)$. We then iteratively pop from $Q$ the pixel with the smallest Manhattan distance to the target pixel~(Item 4), and repeat this procedure until the target is reached~(Item 6). Finally, the recorded parent pointers describe a path from the start to the target pixel, which could be used for splitting the overlapping sentence.

The second case, shown in Fig~\ref{fig:split}~(b), involves pixel-level connection between adjacent components, making it impossible to find a non-destructive separating path. To address this, \textit{Aggressive Split} segments the overlapping text by searching for a division line that minimally disrupts the foreground text region. We firstly divide the bounding box into a sequence of bins along the vertical direction, and then construct a histogram by counting the foreground pixels within each bin, followed by Gaussian smoothing to reduce variability between adjacent bins. The significant local minima of the smoothed histogram are then chosen as division lines, which could segment the multiple-sentence SCC into several single-sentence SCCs. Although this procedure may assign a tiny portion of a SCC to an adjacent one, the affected area is negligible and does not significantly impact subsequent SCC classification.

\subsection{Region-aware Handwriting Descriptor}\label{sec:feature}
After segmenting the document image into sentence-level components in the previous stage, the next step is to design effective features to distinguish between \textit{handwritten} and \textit{printed} text sentences. Visually, handwritten text exhibits irregular strokes and subtle color variations, whereas printed text tends to maintain uniformity in both structure and intensity. Although these two categories are intuitively separable for humans due to their distinct spatial distributions, their global statistics at the SCC level (e.g., mean and standard deviation) show only marginal differences. 
This observation suggests that global descriptors overlook critical positional cues.To overcome this limitation, we propose a \textit{Region-aware Handwriting Descriptor} that explicitly captures spatial and channel-wise variations within the text regions. The proposed RHD consists of two key components: (1)~\textit{Region Division}, which partitions the input image into several subregions to preserve spatial locality; and (2)~\textit{Feature Extraction}, which computes concise statistical and geometric descriptors from each subregion to represent structural and chromatic characteristics of handwritten and printed text.

\subsubsection{Region Division}\label{sec:division}
In this work, we partition the image into several concentric annuli centered at the image center. There are two main reasons for this design: (1)~the concentric annuli preserve coarse positional information through the distance-to-center while remaining compact, and (2)~the horizontal and vertical symmetry of each annulus ensures that per-annulus statistics accurately reflect the spatial distribution of content. Let the image width and height be \(W\) and \(H\), respectively. For any pixel \(p=(p_x,p_y)\), its Euclidean distance to the central pixel \(cp=\bigl(\tfrac{W}{2},\tfrac{H}{2}\bigr)\) can be defined as $D_{p}=\sqrt{(p_x - W/2)^2 + (p_y - H/2)^2}$.

We define \(n+1\) a strictly increasing radial threshold sequence 
$
0 = S_0 < S_1 < \cdots < S_{n} = \sqrt{\frac{W^2}{4} + \frac{H^2}{4}}
$
to partition the image into \(n\) concentric regions \(R_0, R_1, \dots, R_{n-1}\), where $R_i = \{p | S_i < D_{p\text{-}cp} \le S_{i+1} \}$. In this study, we present three methods to decide the radial thresholds, i.e, linear partitioning, square-law partitioning, and log-polar partitioning.  Since $S_0$ and $S_n$ are fixed, we focus our discussion on the intermediate thresholds $S_i$ for $i = 1, 2, \dots, n{-}1$.

\noindent \textbf{Linear Partitioning.} The image is partitioned into $n$ equal-width concentric annuli, i.e, $S_i = ({i}/{n}) S_n$.

\noindent \textbf{Square Partitioning.} To emphasize central details, the radial domain is partitioned with denser thresholds near the origin and sparser ones toward the boundary. This is achieved using an inverse-square progression, i.e., $S_i = S_n / {(n + 1 - i)^2}$.

\textbf{Log-polar Partitioning.} The image is partitioned into $n$ concentric annuli, whose radial boundaries are equally spaced on the logarithmic scale, i.e., $S_i = (S_n)^{i/n}$.

\subsubsection{Feature Extraction}\label{feature}
To keep the method simple and efficient, we extract only a few basic statistical features: mean, sum of squared deviations, area, and aspect ratio.  
Their definitions are as follows:  

\begin{enumerate}
	\item \textbf{Mean}, reflecting a region's overall brightness, is defined as
	\begin{equation}\label{eq:muE}
		\mu_{R}^c = \frac{1}{|R|}\sum_{p \in R} I_c(p),
	\end{equation}
	where $p$ denotes a pixel in region $R$, $|R|$ is the number of pixels in $R$, and $I_c(p)$ is the value of the $c$-th channel at pixel $p$.
	
	\item \textbf{SSD}, characterizing intra-region uniformity and variability, is defined as
	\begin{equation}
		SSD_{R}^c = \sum_{p \in R} \left( I_c(p) - \mu_{R}^c \right)^2.
	\end{equation}
	
	\item \textbf{Area}, measuring the size of a region, is defined as
	\begin{equation}
		A_R = |R|.
	\end{equation}
	
	\item \textbf{Aspect ratio}, describing the geometric shape of a region, is defined as
	\begin{equation}
		\alpha_R = \frac{w_R}{h_R},
	\end{equation}
	where $w_R$ and $h_R$ denote the width and height of region $R$, respectively.
\end{enumerate}

These features capture the principal statistical and shape properties of text regions at low computational cost, serving as a robust foundation for subsequent classification and segmentation. Finally, we construct the region-aware human-writing descriptor $F_{\text{RHD}}$ by concatenating the above four features of each region and imgage channel:	
\begin{equation}
	F_{\text{RHD}} \;=\;
	\bigcup_{R}
	\Bigl(
	\bigcup_{c}
	\{\mu_{{R}}^c,\;\operatorname{SSD}_{{R}}^c\}
	\;\cup\;
	\{A_{R},\;\alpha_{R}\}
	\Bigr).
\end{equation}

With the constructed RHD, we can easily segment the handwritten and printed text via a conventional classifier. Since the text masks of each SCC have already been obtained during the Sentence-level Connected Component Segmentation process, the final handwritten and printed text segmentation can be achieved by simply merging the masks corresponding to the two categories.

\section{Experiments}

\subsection{Experimental Settings}

\begin{figure*}[htb]
	\centering
	\includegraphics[width=0.8\linewidth]{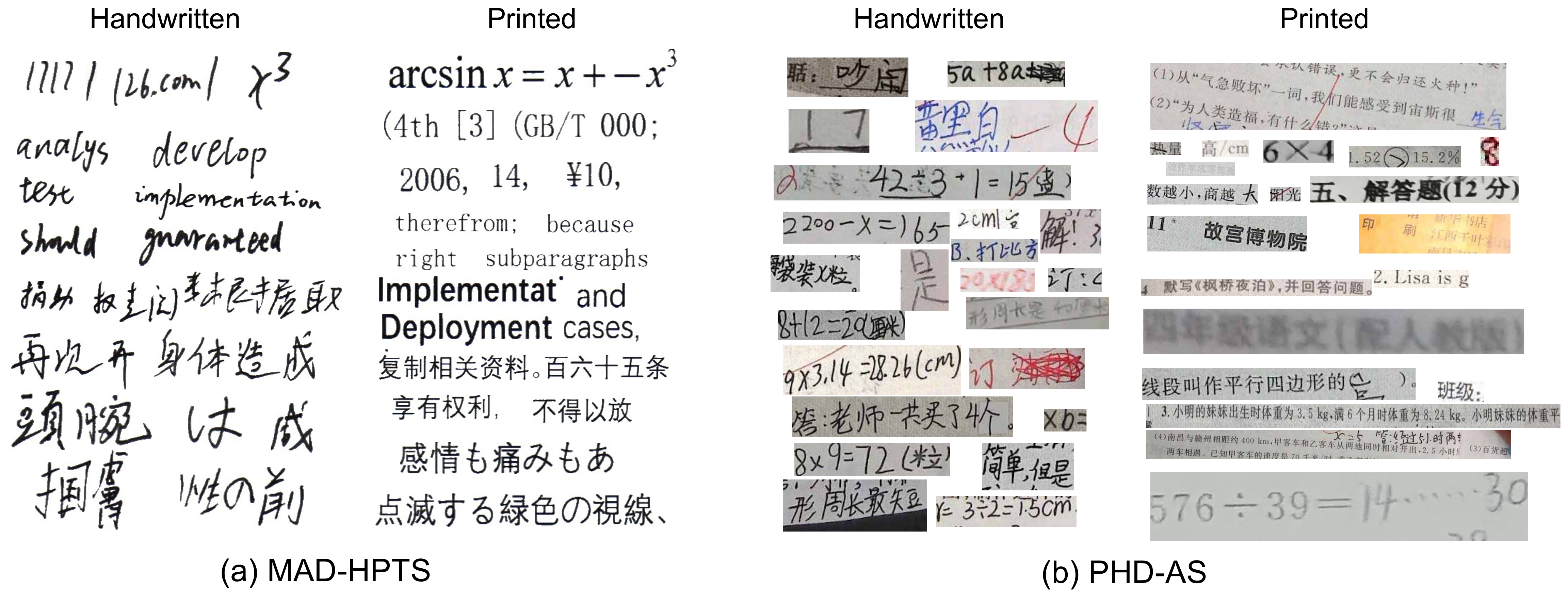}
    \caption{Examples from the two datasets. (a)~MAD-HPTS provides clean and noise-free scans after background removal, covering multiple languages and scripts including English, Chinese, Japanese, and Arabic numerals. (b)~PHD-AS contains real-world images with perspective distortion, motion blur, and background clutter, reflecting challenging acquisition conditions.}
	\label{fig:datasets}
\end{figure*}

\subsubsection{Datasets}
To evaluate the generalization capability of the proposed method, we conduct experiments on two datasets as shown as Fig~\ref{fig:datasets}: the left panel show the self-collected Multilingual Annotated Dataset for Handwritten and Printed Text~(MAD-HPTS) and the right panel show a publicly available benchmark~(PHD-AS),

\noindent \textbf{MAD-HPTS} To rigorously assess the proposed method under diverse scenarios, we constructed a large-scale dataset encompassing variations in handwriting styles, pen types, scanning devices, languages, and document types. Specifically, 200 A4 sheets were printed with text drawn from news articles, contracts, and other sources. Three individuals then added handwritten annotations on these sheets using two different black pens, covering Chinese characters, Arabic numerals, English, and Japanese. The annotated documents were scanned using two distinct devices based on STM32 microcontrollers, each equipped with a stepper motor operating at constant speed to ensure stable and accurate scanning. In total, 200 documents were scanned, resulting in 400 digital images with a resolution of $2592 \times 3666$. Each image was processed by the SCCS preprocessing method to segment the text into SCC blocks, which were subsequently labeled manually to construct a binary classification dataset. This yielded 107{,}830 samples in total, of which 80\% were randomly selected for training and the remaining 20\% were used for testing. The MAD-HPTS dataset will also be made publicly available through the Paddle AiStudio community.

\noindent \textbf{PHD-AS} The PHD-AS dataset is derived from the Printed Handwritten Data released on the Paddle AiStudio community~\cite{paddledataset2022}. It contains 4,800 images in total, with 2,000 printed and 2,000 handwritten samples for training, and 400 printed and 400 handwritten samples for testing. The dataset originates from student assignment scenarios and was captured by camera, with content primarily in Chinese and Arabic numerals. As a binary-classification dataset, PHD-AS aligns naturally with the task studied in this work and serves as a complementary benchmark to MAD-HPTS.

\begin{figure*}[htb]
	\centering
	\includegraphics[width=\linewidth]{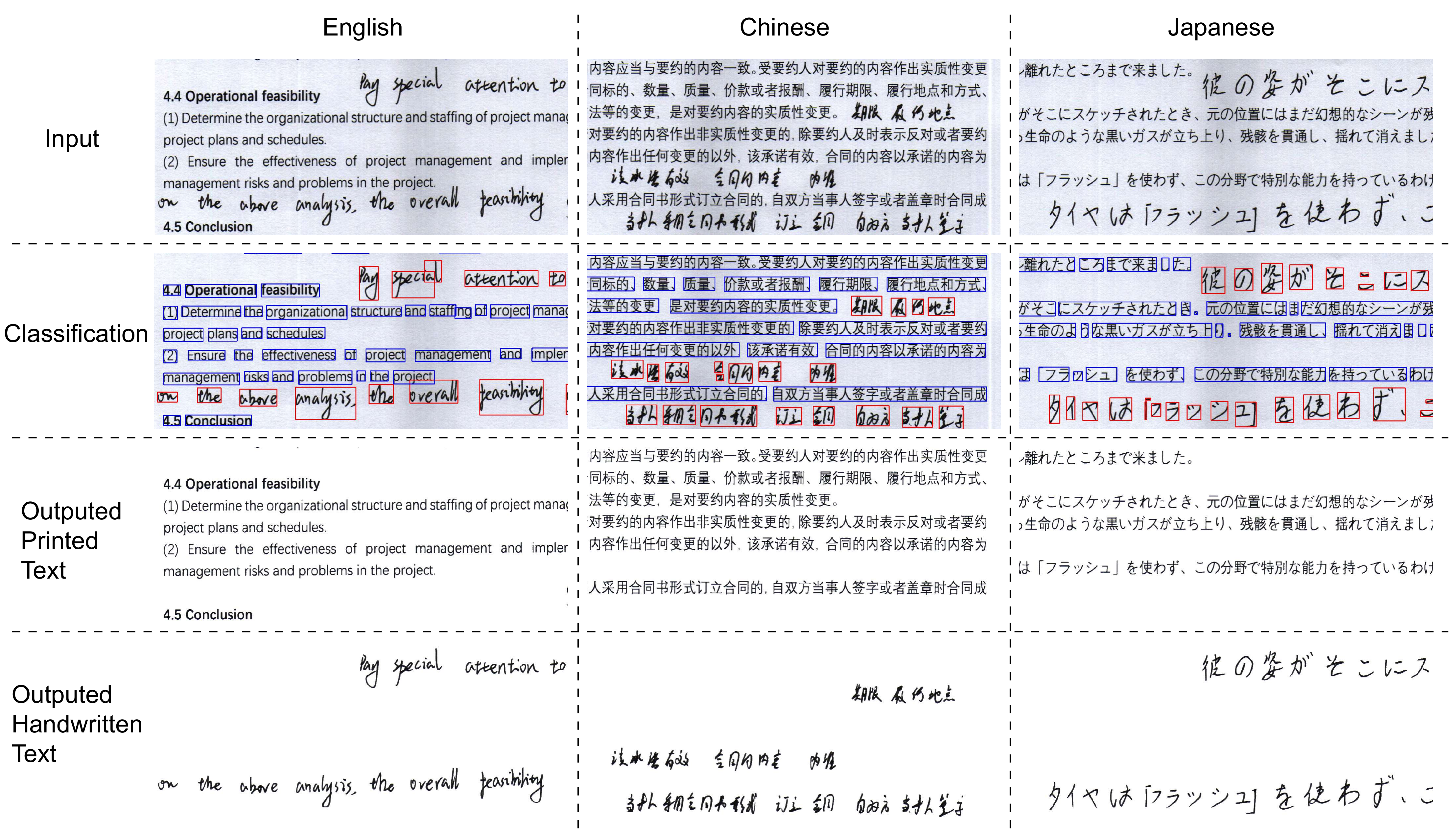}
	\caption{Segmentation results on the MAD-HPTS dataset for English, Chinese, and Japanese documents. From top to bottom: input images, classification (blue boxes for printed text and red for handwritten text), printed text, and handwritten text.}
	\label{fig:sample}
\end{figure*}

\subsubsection{Baselines}
To comprehensively evaluate the effectiveness of the proposed method, we compared it against two categories of approaches: (1)~traditional machine learning-based methods, and (2)~lightweight deep learning-based methods.  
Since handwritten and printed text segmentation pipelines typically consist of preprocessing, feature extraction, and classification modules, direct end-to-end comparisons are often infeasible. To ensure fairness, we standardized the preprocessing and classifier components across all methods, focusing the comparison on feature extraction modules for machine learning-based approaches, and on network architectures for deep learning-based approaches.

\noindent \textbf{Machine Learning-Based Methods.}  
For this category, we implemented six representative handcrafted feature extraction methods:  
(1)~Hu moments~\cite{hu1962visual}, adapted to our task by computing the mean and variance of moment values across all words within each sentence;  
(2)~gray-level co-occurrence matrix (GLCM) texture statistics calculated at $180^{\circ}$ with a pixel distance of 1~\cite{srivastava2020pattern};  
(3)~uniform local binary patterns (LBP), where the 0--255 codes are aggregated into eight histogram bins~\cite{nanni2012survey};  
(4)~gradient-based directional features that capture both magnitude and orientation information~\cite{tamura1978textural};  
(5)~grayscale intensity descriptors proposed by Malakar \textit{et al.}~\cite{malakar2013handwritten}; and  
(6)~combined intensity–structural descriptors introduced by Garlapati \textit{et al.}~\cite{garlapati2017system}.  

To ensure a fair and consistent comparison focusing exclusively on the feature extraction stage, all methods share identical preprocessing procedures and are evaluated using the same random forest classifier with 20 trees. 
This classifier is selected for its simplicity, robustness, computational efficiency, and independence from GPU acceleration, enabling straightforward deployment on a wide range of edge devices.

\noindent \textbf{Deep Learning-Based Methods.}  
Although deep learning-based methods generally achieve high accuracy, their computational cost can limit their applicability to resource-constrained environments. To reflect realistic deployment scenarios on edge devices, we selected two lightweight neural networks as baselines: (1)~LeNet, which processes fixed-size images of $128 \times 256$ pixels~\cite{lecun2002gradient}; and (2)~a fully convolutional network (FCN) capable of handling images with variable sizes~\cite{long2015fully}. 

\subsubsection{Hyperparameters}
The hyperparameters employed in our framework are summarized as follows. 
In the \textit{Component Initialization} stage~(Section~\ref{CharacterLocalization}), the minimum connected component area is set to 10 pixels to remove small noise regions, the spatial density threshold is fixed at 0.02 to filter sparse components, and the aspect ratio is constrained within the range of $[1/35, 35]$ to accommodate both narrow and wide character shapes. 
For the \textit{Region-aware Handwriting Descriptor}~(Section~\ref{feature}), the number of spatial partitions is fixed to 10, and the division is performed using a linear partitioning strategy to ensure uniform coverage of the sentence region.

\subsection{Visualization of Segmentation Results}
To intuitively demonstrate the effectiveness and generalization ability of the proposed method across different languages, we visualize segmentation results on English, Chinese, and Japanese document samples (Fig.~\ref{fig:sample}). The method exhibits consistent performance across languages with distinct character structures—from the intricate strokes of Chinese characters to the simpler shapes of English letters and Japanese kana. It effectively distinguishes between printed and handwritten text even under variations in fonts and handwriting styles. Notably, the segmentation masks preserve fine-grained details, such as individual handwritten strokes, indicating high precision. Although minor inaccuracies occasionally occur in densely overlapping regions, the method overall achieves robust segmentation accuracy. These qualitative observations complement the quantitative analyses presented in the following section, providing further evidence of the proposed approach’s effectiveness.

			

\begin{table}[htb]
    \centering
    \caption{Comparison with machining learning-based methods}
    \resizebox{0.98\columnwidth}{!}{%
        \begin{tabular}{lcccc}
        \toprule
         & \multicolumn{2}{c}{MAD-HPTS} & \multicolumn{2}{c}{PHD-AS} \\
         \cmidrule(lr){2-3} \cmidrule(lr){4-5}
        Methods & Accuracy(\%) & Runtime(s) & Accuracy(\%) & Runtime(s)\\
        \midrule
        Hu~\cite{hu1962visual} & 86.7 & 11.6 & 71.3 & 1.1\\
        GLCM~\cite{srivastava2020pattern} & 89.1 & 27.5 & 64.0 & 1.7\\
        LBP~\cite{nanni2012survey} & 87.3 & 12.8 & 70.6 & 1.1\\
        Malakar et al.~\cite{malakar2013handwritten} & 90.0 & 11.9 & 66.0 & 0.9\\
        Direction~\cite{tamura1978textural} & 93.5 & 12.5 & 74.0 & 1.1\\
        Garlapati et al.~\cite{garlapati2017system} & 94.9 & 10.7 & 82.2 & 1.0\\
        RHD~(Ours) & \textbf{96.9} & \textbf{8.7} & \textbf{83.8} & \textbf{0.4}\\
        \bottomrule
        \end{tabular}%
    }
    \label{tab:experiment1}
\end{table}

\subsection{Comparison with Machine Learning-Based Methods}\label{sec:experience1}
To systematically assess the robustness and computational efficiency of the proposed RHD method against established feature extraction approaches, we conducted comprehensive experiments on two benchmark datasets, MAD-HPTS and PHD-AS. Both datasets contain precisely annotated text regions, ensuring that the evaluation isolates the performance of feature extraction methods. RHD was compared with six well-established baselines~\cite{hu1962visual,srivastava2020pattern,nanni2012survey,tamura1978textural,malakar2013handwritten,garlapati2017system}, and performance was evaluated using multiple metrics. For each metric, the median values over repeated trials were reported to ensure result stability and statistical representativeness. 

The experimental results summarized in Table~\ref{tab:experiment1} clearly demonstrate that RHD consistently achieves the best performance across all evaluation metrics on both MAD-HPTS and PHD-AS. 
In addition to superior accuracy, RHD also maintains the lowest computational overhead among all evaluated methods, underscoring its practical efficiency. Notably, although PHD-AS is captured under real-world conditions with perspective distortion, motion blur, and irregular text alignment, RHD still attains the highest performance under the challenging dataset. This consistent superiority in both accuracy and computational cost further substantiates the effectiveness and robustness of the proposed RHD in diverse, real-world text segmentation scenarios.

			

\begin{table}[htb]
    \centering
    \caption{Comparison with deep learning-based methods}
    \resizebox{0.98\columnwidth}{!}{%
        \begin{tabular}{lcccc}
        \toprule
         & \multicolumn{2}{c}{MAD-HPTS} & \multicolumn{2}{c}{PHD-AS} \\
         \cmidrule(lr){2-3} \cmidrule(lr){4-5}
        Methods & Accuracy(\%) & Runtime(s) & Accuracy(\%) & Runtime(s)\\
        \midrule

        CNN~\cite{lecun2002gradient} & 73.7 & 229.3 & 71.3 & 8.7\\
		FCN~\cite{long2015fully} & \textbf{98.3} & 67.5 & \textbf{91.5} & 6.3 \\		
		RHD~(Ours) & 96.9 & \textbf{8.7} & 83.8 & \textbf{0.4} \\
       
        \bottomrule
        \end{tabular}%
    }
    \label{tab:nn}
\end{table}

\subsection{Comparison with Deep Learning-Based Methods}
To further evaluate the practical efficiency of the proposed RHD method relative to more powerful neural networks, we conducted experiments comparing two representative architectures on MAD-HPTS and PHD-AS. All experiments were performed on the same CPU-only workstation to ensure a fair and consistent comparison. As summarized in Table~\ref{tab:nn}, the proposed RHD method runs over eight times faster than LeNet on MAD-HPTS and more than 15 times faster on PHD-AS, demonstrating significant practical advantages for deployment in mobile or other low-power edge scenarios. Although RHD is slightly less accurate than FCN, achieving only 1.4 percentage points lower accuracy on MAD-HPTS, the difference is modest. 
This small gap is likely due to the relatively simple feature distributions in the scanner-acquired MAD-HPTS dataset, which can be effectively captured by handcrafted region-aware descriptors. 
On the more challenging, camera-acquired PHD-AS dataset, the accuracy difference is larger, reflecting the greater benefits of end-to-end feature learning under irregular text layouts. Although FCN is more accurate, its much longer inference time limits practicality on resource-constrained devices. In real-world deployments, a minor accuracy trade-off for an order-of-magnitude speed gain often represents the most pragmatic choice, highlighting the effectiveness of the proposed RHD approach for efficient, cross-dataset handwritten and printed text classification.

\begin{table}[htb]
    \centering
    \caption{Performance comparison on RK3576 }
    \resizebox{0.75\columnwidth}{!}{%
        \begin{tabular}{lcc} 
			\toprule
			
			Methods  & Accuracy(\%) & Runtime(s) \\ 
			\midrule		
			LBP~\cite{nanni2012survey} & 85.1 & 12.3 \\
			Hu~\cite{hu1962visual} & 85.2 & 21.2\\
			GLCM~\cite{srivastava2020pattern} & 87.7 & 98.3\\
			Malakar et al.~\cite{malakar2013handwritten} & 88.4 & 20.9 \\
			Direction~\cite{tamura1978textural} & 92.8 & 13.2 \\
			Garlapati et al.~\cite{garlapati2017system} & 94.3 & 19.8 \\
			RHD~(Ours) & \textbf{95.9} & \textbf{10.8} \\
			
			\bottomrule
		\end{tabular}
    }
    \label{tab:experiment4}
\end{table}

\subsection{Performance on Edge Devices}
To further evaluate the practical applicability and deployment efficiency of the proposed method on resource-constrained edge devices, we deployed all methods on the RK3576 platform. 
Experiments were conducted using 3,000 randomly selected handwritten and 3,000 printed samples from the MAD-HPTS dataset, resulting in a total of 6,000 test images. For each method, both the overall classification accuracy and the total inference time across all images were recorded. Except for the hardware platform and the number of test samples, the experimental setup remained identical to that described in Section~\ref{sec:experience1}.

Experimental results, summarized in Table~\ref{tab:experiment4}, show that RHD continues to achieve the highest accuracy among all tested methods. The slight decrease in accuracy compared to previous experiments is likely due to the balanced selection of handwritten and printed testing samples. Despite this, RHD still clearly outperforms all alternative approaches. Moreover, on the edge device, RHD maintains the lowest inference time, confirming its efficiency and suitability for real-world deployment on resource-constrained platforms.

			
			
			

%
\begin{figure*}[htb]
	\centering
	\includegraphics[width=\linewidth]{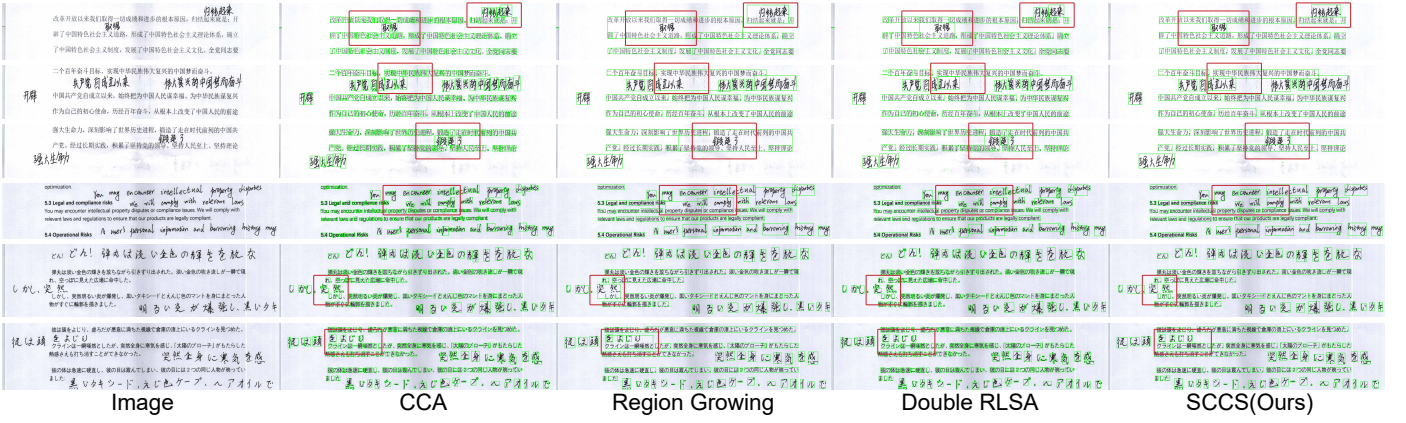}
	\caption{The figure qualitative comparison of four text-segmentation algorithms on seven samples. For each row, columns from left to right show: original image, CCA method~\cite{rosenfeld1966connected}, region-growing algorithm~\cite{shirdhonkar2010discrimination}, double-RLSA algorithm~\cite{belaid2013handwritten}, and the proposed SCCS method. Red bounding boxes highlight regions where the differences between methods are most visually distinguishable.}
	\label{fig:segmentation}
\end{figure*}

\subsection{Performance with Different Classifiers}
To further evaluate the generality of the proposed RHD features, we compared them with the features of Garlapati et al.~\cite{garlapati2017system}, which achieved the second-best performance in the previous machine learning-based comparison. 
Specifically, both RHD and Garlapati’s features were used as inputs to several classical machine learning classifiers, including Decision Tree, Gradient Boosting, RBF SVM, and Random Forest~\cite{cortes1995support,breiman1984cart,friedman1999stochastic,breiman2001random}. 

The results, summarized in Table~\ref{tab:classifiers}, show that RHD features consistently outperform Garlapati’s features across all classifiers, demonstrating stronger discriminative capability, especially for complex decision boundaries. 
Notably, the combination of RHD features with Random Forest achieves the highest performance, with an accuracy of 96.9\% and an F1-score of 95.5\%, surpassing the results obtained by Garlapati’s features with any classifier.

\begin{table}[htb]
    \centering
    \caption{Accuracy~(\%) performance comparison of different classifiers.}
    \resizebox{0.85\columnwidth}{!}{
        \begin{tabular}{lcc}
        \toprule
        
        Classifiers & Garlapati's et. al~\cite{garlapati2017system}  & RHD~(Our)\\
        \midrule
        Decision Tree & 93.8 & \textbf{94.1}\\
        Gradient Boosting & 93.9 & \textbf{94.3} \\
        RBF SVM & 93.3 & \textbf{94.9}\\
        Random Forest & 94.9 & \textbf{96.9} \\

        \bottomrule
        \end{tabular}%
    }
    \label{tab:classifiers}
\end{table}

\subsection{Performance of SCC Segmentation}
Due to the lack of standardized quantitative metrics suitable for evaluating text segmentation quality in our context, and considering that segmentation is not the most critical module of the system, we perform a qualitative analysis to assess segmentation performance. Over-segmented SCCs do not necessarily lead to errors; for instance, a single character may be split into multiple SCCs, but these can still be correctly classified in subsequent stages. Specifically, we visually analyzed and compared the results obtained from four segmentation algorithms: the connected component analysis~(CCA)~\cite{rosenfeld1966connected}, the region growing algorithm~\cite{shirdhonkar2010discrimination}, the double RLSA algorithm~\cite{belaid2013handwritten}, and our proposed SCCS method.

The comparison results are shown in Fig.~\ref{fig:segmentation}. 
The evaluation images were carefully selected to include complex layouts in Chinese, English, and Japanese, enabling a comprehensive visual comparison of the four segmentation algorithms. The CCA method produces clean character-level segmentation across all languages, but often over-segments sentences into individual radicals or strokes, which could negatively impact subsequent feature extraction, particularly for Chinese and Japanese characters. In contrast, the region growing algorithm tends to under-segment by merging content from multiple sentences into a huge region, frequently grouping overlapping text lines and intersecting bounding boxes. The double RLSA algorithm alleviates some under-segmentation, yet it still fails to connect multiple words within a sentence, particularly in densely written regions of Japanese handwriting. By comparison, the proposed SCCS method yields the most consistent and coherent segmentation results across all three languages. Printed and handwritten text are cleanly separated, overlapping lines remain distinct, and even small handwritten insertions between printed lines are accurately isolated, demonstrating the superior performance and robustness of SCCS in handling complex multilingual layouts.

\subsection{Enhancing Existing Methods with Region-Aware Feature Design Principle}
To validate the effectiveness and generality of the proposed region-aware feature design principle, we applied a simple concentric-region partitioning strategy to five existing feature extraction methods: GLCM~\cite{srivastava2020pattern}, LBP~\cite{nanni2012survey}, Direction~\cite{tamura1978textural}, Malakar et al.'s grayscale features~\cite{malakar2013handwritten}, and Garlapati et al.'s intensity-structural features~\cite{garlapati2017system}. Each image was divided into three concentric regions, and features were extracted separately from each region and then concatenated to form a single feature vector for classification. 

\begin{figure}[htb]
	\centering
	\includegraphics[width=\linewidth]{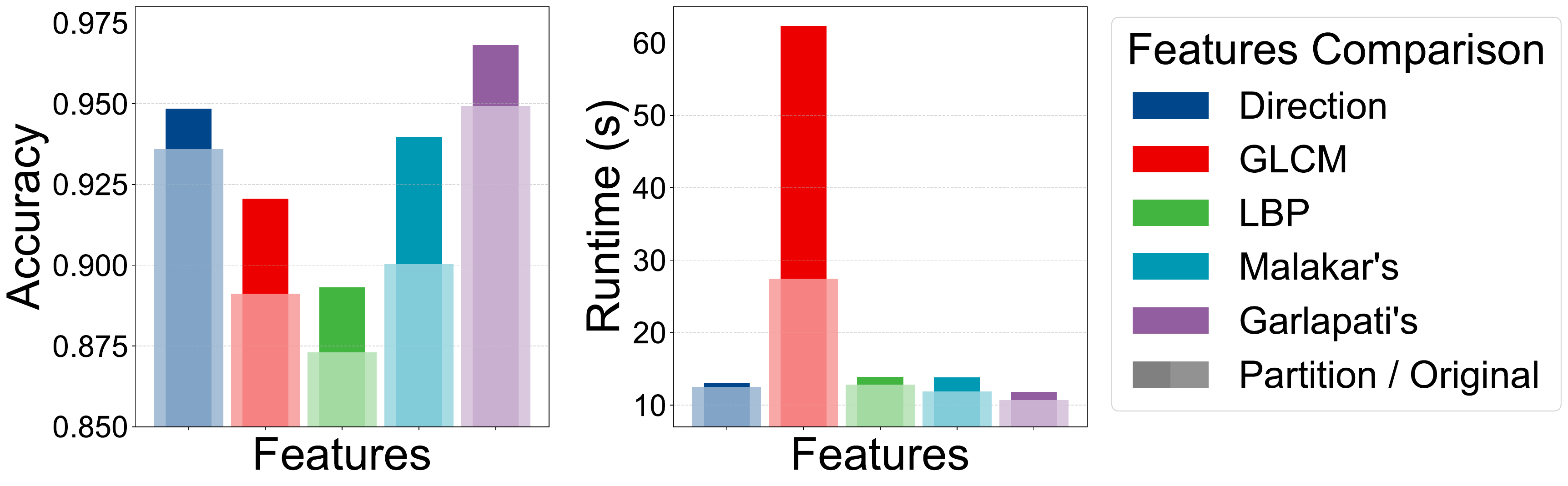}
	\caption{The figure show the existing methods enhanced by region-aware feature design principle. For each feature, the solid bars show the original accuracy or runtime, while the semi-transparent bars indicate the accuracy or runtime after partitioning.}
	\label{fig:expansion}
\end{figure}

The comparative results are shown in Fig.~\ref{fig:expansion}. 
All five methods benefit from region-aware partitioning, achieving noticeable accuracy improvements with minimal additional computational cost. Notably, Garlapati et al.'s method achieved over 96\% accuracy after partitioning, with only a 10.3\% increase in processing time. For most texture-based descriptors, the partitioning introduces negligible overhead. In contrast, GLCM requires computing statistics on a $256\times256$ co-occurrence matrix for each region, causing its computational cost to grow nearly linearly with the number of partitions due to repeated matrix computations, which reduces the cost-effectiveness of this optimization. Overall, these results demonstrate that the region-aware feature design principle can consistently enhance traditional feature extractors, highlighting the generality and practical utility of the proposed RHD approach.

\subsection{Comparison of Region Partition Strategies}
The performance of the RHD method is influenced by two key hyper-parameters: the number of partition regions and the partitioning strategy. To systematically investigate their impact, we conducted two sets of experiments, focusing separately on the number of regions and the choice of partitioning strategy.

\subsubsection{Number of Partition Region}
\begin{figure}[htb]
	\centering
	\includegraphics[width=\linewidth]{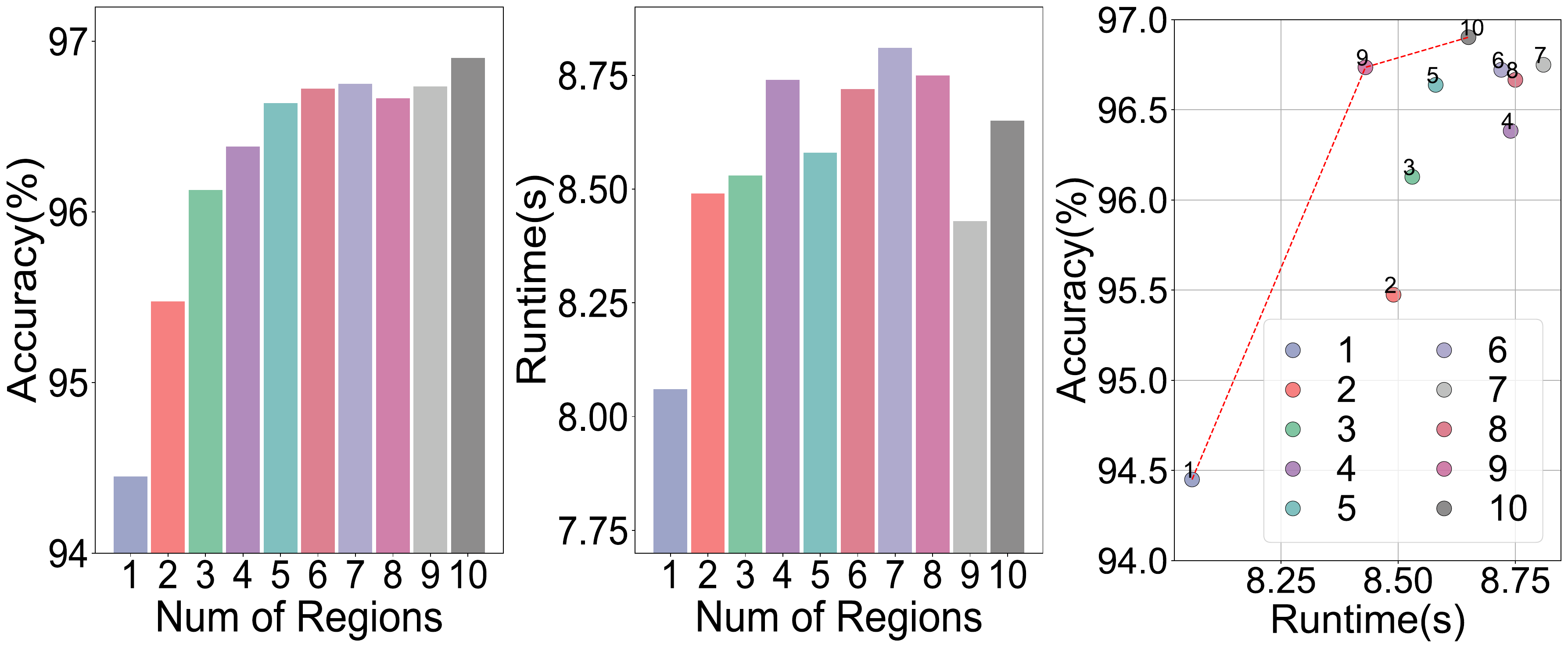}
	\caption{The left figure illustrates classification accuracy as a function of the number of regions;  the middle figure shows runtime versus region count;  and the right figure depicts the Pareto front between accuracy and region count.}
	\label{fig:numofregion}
\end{figure}

To analyze the impact of partition granularity, we fixed the partition strategy to linear and varied the number of regions from one to ten. For each configuration, we measured both classification accuracy and inference time. The results, as shown in Fig~\ref{fig:numofregion}, indicate that increasing the number of regions generally leads to higher accuracy, as finer divisions capture more local spatial variations within the text regions. Importantly, the improvement in accuracy is accompanied by only a modest increase in computational cost. Runtime fluctuations observed for higher region counts are primarily due to measurement noise and the random variation in the structure of the random forest classifier. Further analysis indicates a clear Pareto frontier in the number of partition regions, highlighting a trade-off between accuracy and inference time. This demonstrates the flexibility of the proposed method, allowing the number of regions to be adjusted according to specific accuracy–time requirements in practical applications.

\subsubsection{Partition Strategy}
To compare the differences among partitioning strategies, we fixed the number of regions to three and systematically compared three distinct partitioning schemes: linear, square, and log-polar. Each scheme divides the image into regions with different shapes and spatial distributions, potentially affecting the extraction of local features and, consequently, the overall classification performance.

As shown in Table~\ref{tab:experiment5}, the linear partition strategy achieves the highest accuracy. Compared with the results in Table~\ref{tab:experiment1}, all three partitioning schemes outperform the other compared methods, indicating that regional partitioning itself contributes positively to feature discrimination. Furthermore, with only a small number of regions, the consistent improvements demonstrate the robustness and generalization capability of the proposed method.

		
		

\begin{table}[htb]
	\centering
	\footnotesize
	\caption{Performance comparison of different partition methods}
	\begin{tabular}{lc} 
		\toprule
		
		Partition methods  & Accuracy(\%) \\ 
		\midrule
		Square & 95.8  \\
		Log-polar & 96.0 \\
		Linear & \textbf{96.1} \\
		
		\bottomrule
	\end{tabular}
	\label{tab:experiment5}
\end{table}

\section{ Conclusion}

In this paper, we present the SCCS algorithm for text segmentation and the RHD feature extraction method for distinguishing handwritten from printed characters. On the MAD-HPTS dataset, RHD boosts accuracy by about 2\% over the historical method of Garlapati et al. and trims per-sample runtime by roughly 19\%. Experimental results demonstrate that our RHD approach consistently achieves the best performance across all evaluation metrics on both the MAD-HPTS and PHD-AS while maintaining minimal computational overhead. When compared with neural-network baselines, RHD’s accuracy on MAD-HPTS is only marginally lower, yet its inference time is less than one-eighth of theirs, yielding a substantial efficiency advantage. We further analyse the impact of region division strategies and region counts, showing that even straightforward partitioning schemes can deliver substantial accuracy improvements with little additional runtime. We hope this work will inspire further research into partition-based feature extraction to optimise performance and reveal deeper patterns in document analysis.

\section*{Acknowledgements}
This work was supported by the Natural Science Foundation of Sichuan Province, China under Grant No. 24NSFSC2722, the National Natural Science Foundation of China under Grant No. 62406043, the Open Project Fund of the Key Laboratory of Digital Innovation of Tianfu Culture under Grant No. TFWH-2025-15, and Transformation Foundation of Tianfu Jincheng Laboratory  under Grant No. 2025ZH013.

\bibliography{reference}

\end{document}